\documentclass[conference,twoside]{csce}

\usepackage[hmargin=.75in,vmargin=1in]{geometry}
\usepackage[american]{babel}
\usepackage[T1]{fontenc}
\usepackage{times}
\usepackage{caption}
\usepackage{subfig}
\usepackage{algorithm,algpseudocode}
\usepackage{algorithmicx}
\usepackage{fancyhdr}
\usepackage{textcomp}
\usepackage{epsfig,graphicx}
\usepackage{xcolor}
\usepackage{amsfonts,amsmath,amssymb}
\usepackage{fixltx2e} % Fixing numbering problem when using figure/table* 
\usepackage{booktabs}

\renewcommand{\thispagestyle}[2]{} 
\title{\bf Machine Learning: A Dark Side of Cancer Computing}           %%%% Replace with your title.

\author{
{\bfseries Ripon Patgiri, Sabuzima Nayak, Tanya Akutota, and Bishal Paul}\\
National Institute of Technology Silchar, Assam, India-788010\\
}

\begin{document}

\maketitle                        %%%% To set Title and Author names.

\begin{abstract}%%%% Replace with your abstract.
Cancer analysis and prediction is the utmost important research field for well-being of humankind. The Cancer data are analyzed and predicted using machine learning algorithms. Most of the researcher claims the accuracy of the predicted results within 99\%. However, we show that machine learning algorithms can easily predict with an accuracy of 100\% on Wisconsin Diagnostic Breast Cancer dataset. We show that the method of gaining accuracy is an unethical approach that we can easily mislead the algorithms. In this paper, we exploit the weakness of Machine Learning algorithms. We perform extensive experiments for the correctness of our results to exploit the weakness of machine learning algorithms. The methods are rigorously evaluated to validate our claim. In addition, this paper focuses on correctness of accuracy. This paper report three key outcomes of the experiments, namely, correctness of accuracies, significance of minimum accuracy, and correctness of machine learning algorithms.  
\end{abstract}

\vspace{1em}

%%%%%%%%%%%%%%%%%%%%%%%%%%%%%%%%%%%%%%%%%%%%%%%%%%%%%%%%%%%%

\section{Introduction}
\begin{figure*}[ht]
\centering
\includegraphics[width=0.9\textwidth]{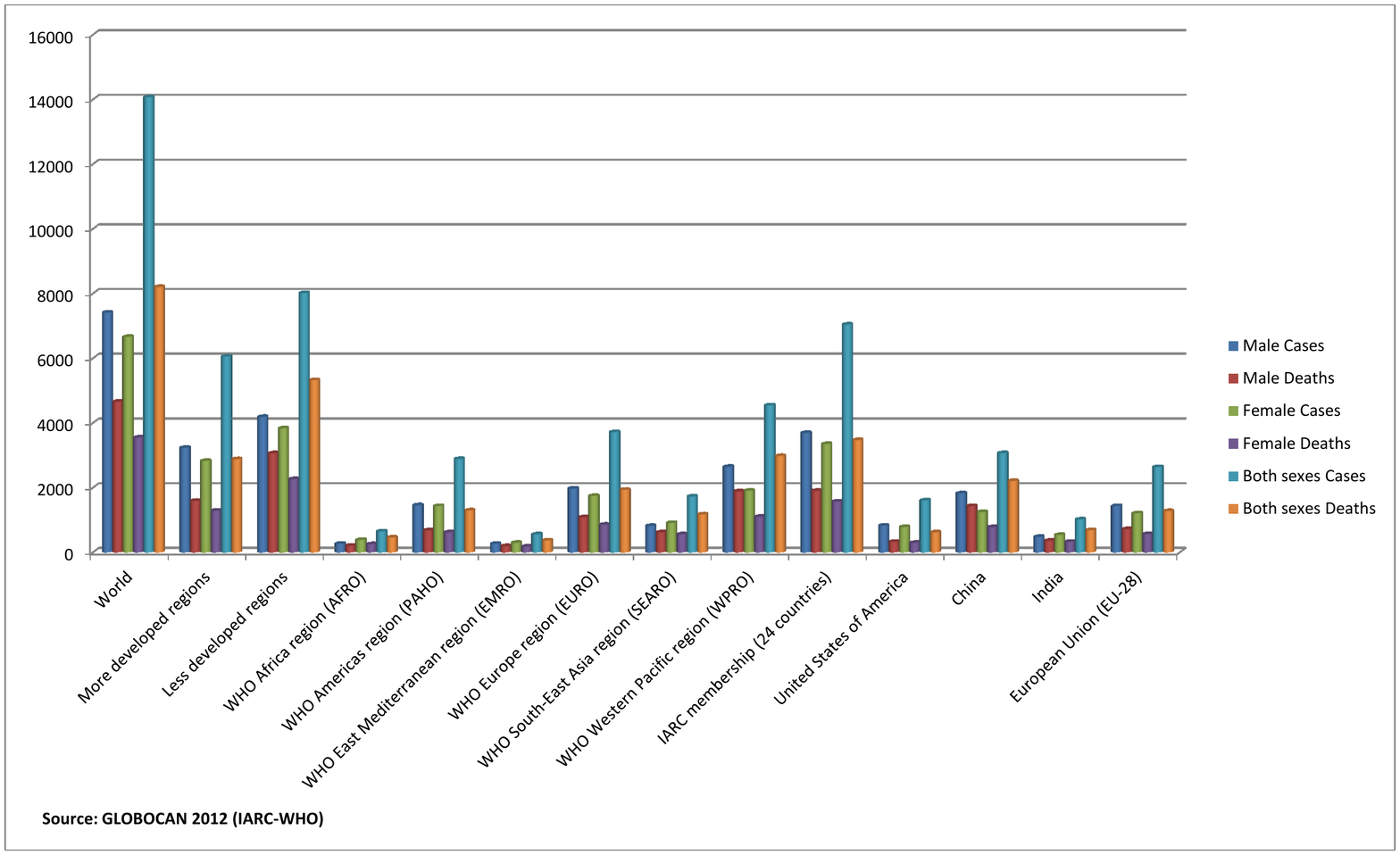}
\caption{Worldwide report of Cancer cases and death \cite{Glob}.}
\label{w}
\end{figure*}

The Cancer, took many lives, and still people are surrendering their lives in front of Cancer. The unpleasant truth is that there is no permanent solution for Cancer till date. However, the Scientists are still trying their level best to save many lives and they are successful too. There are many controversies on "whether a Cancer is a disease or not". Many scientists claim that the Cancer is an unwanted cell behavior due to some mutation. The Scientists believe that the reason of being a Cancer victim may be a high body mass index, low fruit and vegetable intake, lack of physical activity, tobacco use, and alcohol use \cite{WHO}. The definite reason for Cancer is yet to be reported. Many cancer victims could not survive. The cancer mortality is presented in Figure~\ref{w}. However, the modern technology is helping in saving lives of human being from Cancer. For instance, machine learning. The machine learning algorithm plays a vital role in Cancer Computing. The machine learning algorithms are used to analyze the probable presence of Cancer. 

The machine learning algorithms are modified to achieve better accuracy for many purposes, and researchers are developing modern techniques to analyze the Cancer.  Chen et al. \cite{Chen2014} reported accuracy of 83.0\% in lung cancer using Artificial Neural network (ANN) with 440 samples. Xu et al. \cite{xu} reported an accuracy of 97\% in breast cancer using Support Vector Machine (SVM) with 295 sample size. Exarchos et al. \cite{Exa} reported 100\% accuracy in Oral squamous cell carcinoma (OSCC) using their proposed method.  Ahmad et al. \cite{Ahmad} compares three machine learning algorithms on breast cancer, namely, Decision Tree (DT), ANN, and SVM. DT, ANN, and SVM gives an accuracy of 93.6\%, 94.7\%, and 95.7\% respectively using 547 samples.

From the above research results, some research questions (RQ) arise which are given below-
\begin{description}
\item[RQ1:] How can we achieve 100\% accuracy, using machine learning algorithms in prediction of Cancer? Is it ethical?
\item[RQ2:] Can a machine learning algorithm be misled?
\item[RQ3:] Why does researcher emphasize on enhancing the maximum accuracy? Is it really necessary for Cancer prediction?
\item[RQ4:] When can we believe or deploy the proposed machine learning algorithm of a researcher based on their research result?
\end{description}
The research questions are really difficult to answer. However, we critically analyze the research result based on our research questions. RQ1 introduces another dimension to think on machine learning algorithms. It forces to think on ethical and unethical way of gaining accuracy. Similarly, RQ2 also gives indications on the possible misleading of machine learning algorithm. Most importantly, the RQ3 creates a controversial thoughts on maximum and minimum accuracy. Interestingly, RQ4 emphasizes to think about the reliability of the research result with machine learning algorithms. Thus, these four RQs forces to rethink on the machine learning algorithms in dangerous diseases, like Cancer.

We neither present any propose a model nor have any intention to increase the accuracy of the machine learning algorithm. On the contrary, we exploit the behavior of machine learning algorithms and its consequences. In this paper, we present following key points-
\begin{itemize}
\item Experimentation results using WDBC dataset.
\item Experimentation results using doubling the WDBC dataset.
\item Behavior of machine learning algorithms.
\item Significance of minimum accuracy in dangerous diseases.
\item Discloses unethical way of misleading the algorithms.
\end{itemize}

The paper is organized as follows- Section~\ref{bg} discusses on various machine learning approaches to predict Cancer. Section~\ref{dm} provides data and methods to perform experiments. Section~\ref{er} discusses in-depth on the results of our experimentations. Section~\ref{dis} discusses various aspects of machine learning algorithms. And finally, Section~\ref{con} concludes the paper.

\section{Background}
\label{bg}
With the large amounts of cancer data available to work with, machine learning methods have become a de-facto standard of predicting cancer. Machine learning algorithms uncover and identify patterns and extract relationships among the complex data. Prediction accuracy depends on different parameters like patient's age, stage of cancer, medical history, lifestyle, food habits, gender, region based factors, diagnosis histopathology, etc. \cite{Jhajharia}. The accuracy of cancer prediction outcome has significantly improved by 15\%-20\% in the last years, with the application of ML techniques \cite{Kourou}. Kourou et al.\cite{Kourou} compares some of these techniques for breast cancer prediction,  namely Neural Network (NN), Bayesian Network (BN), SVMs and DTs. Barracliffe et al. \cite{Luke} achieve of an 83.6\% accuracy on breast cancer using SVM. The age of the female breast cancer victims are from 28 to 85 years. Surprisingly, Kesler et al. \cite{Kesler} achieves 100\% accuracy using Random Forest model in breast cancer. In addition, there are numerous research on various cancer types. Delen et al \cite{Delen} compares the results of decision tree and neural networks applied to the SEER dataset, with the C5 decision tree having a 93.6\%accuracy compared to Neural Networks, with a 91.2\%. Hamsagayathri and Sampath \cite{Hamsagayathri} proposed a priority-based decision tree which achieves 98.5\% accuracy. Nguyen et al. \cite{Nguyen} have presented the application of random forest combined with feature extraction applied to the diagnosis and prognosis of breast cancer. Their testing accuracy averaged as one of the highest, around 99.8\%. Data set is first N-fold cross validated, estimation of Bayesian probability is done, followed by estimation of feature ranking and value. 

\subsection{Importance of accuracy}
Figure~\ref{acc1}, and ~\ref{acc2} shows the accuracy statistics of hundred rounds on WDBC dataset. The statistics show that accuracy never remains same for the same input data without changing anything. Most of the article reports the maximum accuracy of their algorithm. The maximum accuracy changes time to time. Early conclusion of maximum accuracy with a few runs is incorrect. Because, the accuracy always changes with time. In addition, it is unreliable calculation of the mean accuracy after 5 or 10 runs of their proposed model. A series of experiments are conducted to evaluate the algorithms' performance in terms of mean accuracy. Moreover, the minimum accuracy with two or three experiments to evaluate the algorithms is incorrect. Because, sometimes the algorithms give poor performance in minimum accuracy.

\subsubsection{Significance of minimum accuracy}
It is not required to report the maximum accuracy of a Cancer prediction using a proposed system. The maximum accuracy is in the best case scenario and the algorithm cannot perform beyond that accuracy. It is not so important to become a successful machine learning algorithm in a Cancer or other life threatening diseases. The mean accuracy is an important parameter to report the proposed machine learning algorithms. It is very useful to study cancer disease and to benchmark with other existing algorithm. Now, the most important is the minimum accuracy. The minimum accuracy is the worst case scenario. The worst case scenario dictates the strength of the proposed machine learning algorithm. If people can prove that the proposed system cannot go beyond the minimum benchmark, then the proposed algorithm is reliable. We cannot rely on the maximum accuracy report in life threatening disease. Benchmarking using minimum accuracy gives us more impact than benchmarking using maximum accuracy. Let, Method X and Method Y give result of $m\%$ and $n\%$ in maximum accuracy respectively where $m>n$. Thus, the Method X is better than Method Y. Let, Method X and Method Y give result of $p\%$ and $q\%$ in minimum accuracy where $p<q$. In this case, the Method Y is better than Method X. In a life threatening disease, we cannot rely on Method X, since its minimum accuracy is lower than Method Y.

\section{Data and Methods}
\label{dm}

\begin{table}[ht]
\caption{Parameters of Wisconsin Diagnostic Breast Cancer dataset}
\begin{tabular}{p{3cm}p{4cm}}
\hline
\textbf{Name} & \textbf{Description} \\
\hline
ID & Identity of the patients\\
Diagnosis & M- Malignant and B- Benign\\
Number of features & 32 \\
Number of patients & 569 \\
\hline
\end{tabular}
\end{table}

\begin{figure}[ht]
\centering
\includegraphics[width=0.45\textwidth]{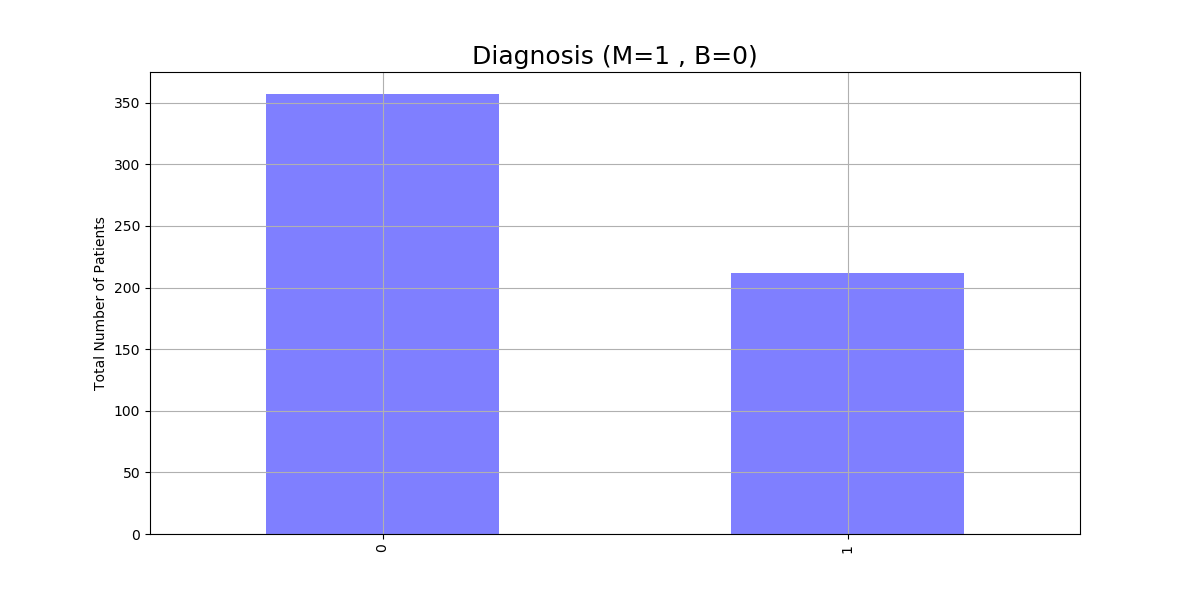}
\caption{Number patients with malignant and benign cancer.}
\label{MB}
\end{figure}

We have experimented on a well-known standard dataset, Wisconsin Diagnostic Breast Cancer (WDBC) dataset. The dataset contains malignant and benign patients. The dataset consists of 569 patients reports. Figure~\ref{MB} depicts the data of malignant patients and benign patient. The dataset contains 212 malignant and 357 benign cancers.

\begin{figure*}[ht]
\subfloat[][Diagnosis vs perimeter mean]{\includegraphics[width=0.3\textwidth]{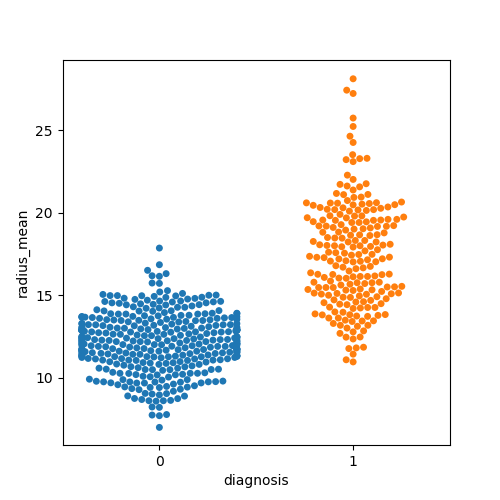}}
\subfloat[][Diagnosis vs texture mean]{\includegraphics[width=0.3\textwidth]{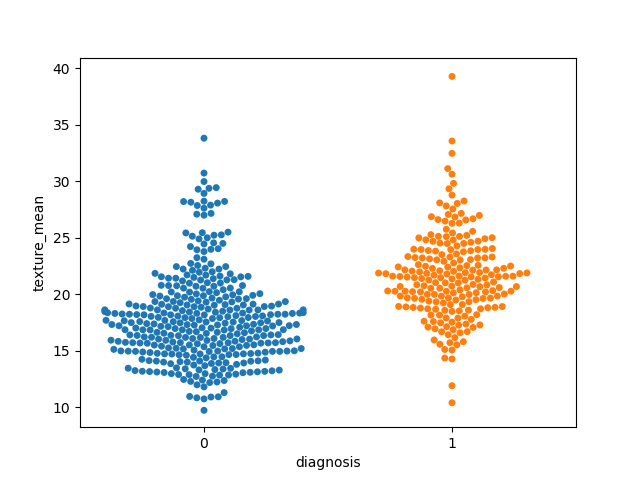}}
\subfloat[][Diagnosis vs perimeter mean]{\includegraphics[width=0.3\textwidth]{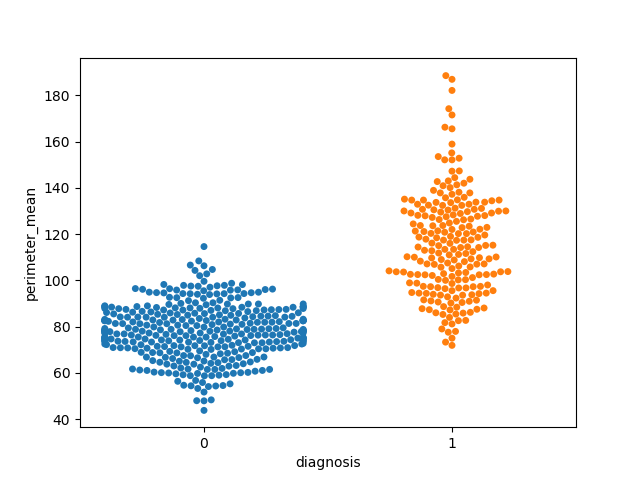}}\\
\subfloat[][Diagnosis vs area mean]{\includegraphics[width=0.3\textwidth]{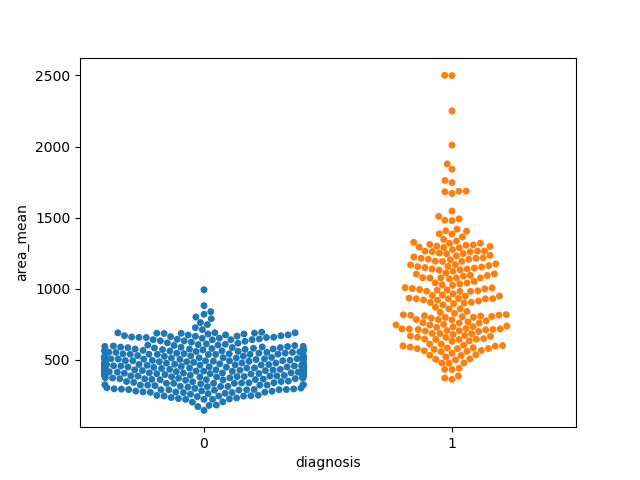}}
\subfloat[][Diagnosis vs smoothness mean]{\includegraphics[width=0.3\textwidth]{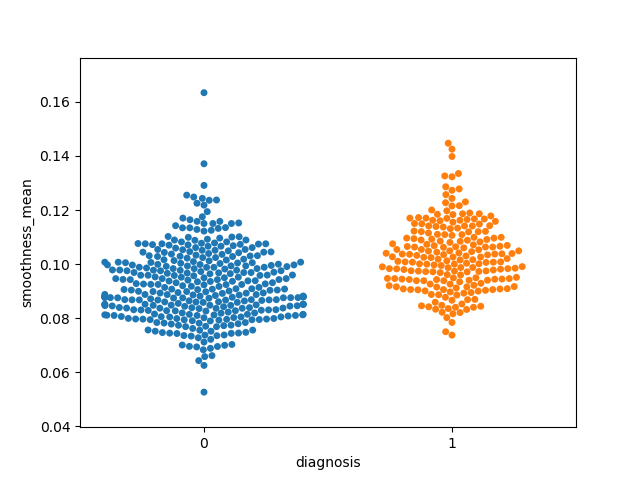}}
\subfloat[][Diagnosis vs compactness mean]{\includegraphics[width=0.3\textwidth]{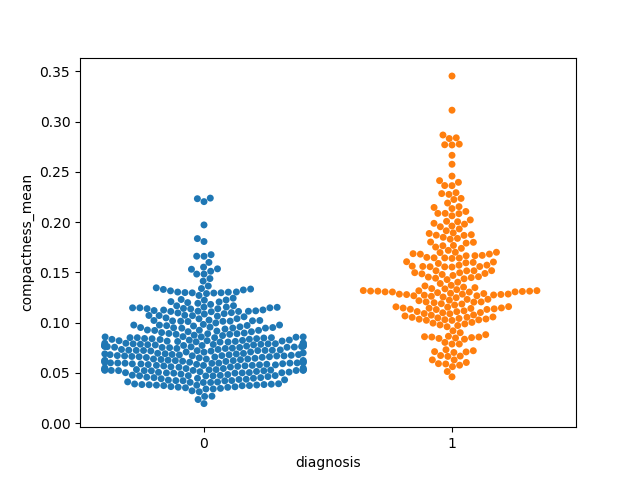}}\\
\subfloat[][Diagnosis vs Concavity mean]{\includegraphics[width=0.3\textwidth]{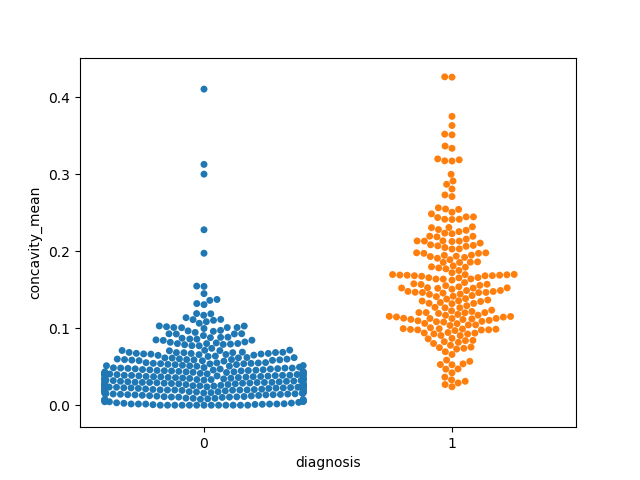}}
\subfloat[][Diagnosis vs concave point mean]{\includegraphics[width=0.3\textwidth]{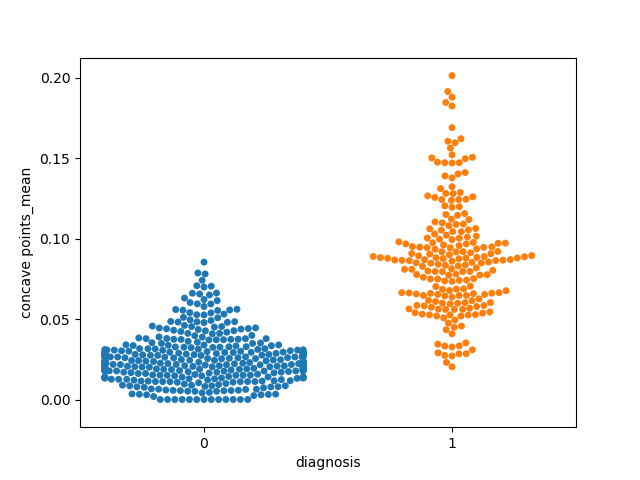}}
\subfloat[][Diagnosis vs symmetry mean]{\includegraphics[width=0.3\textwidth]{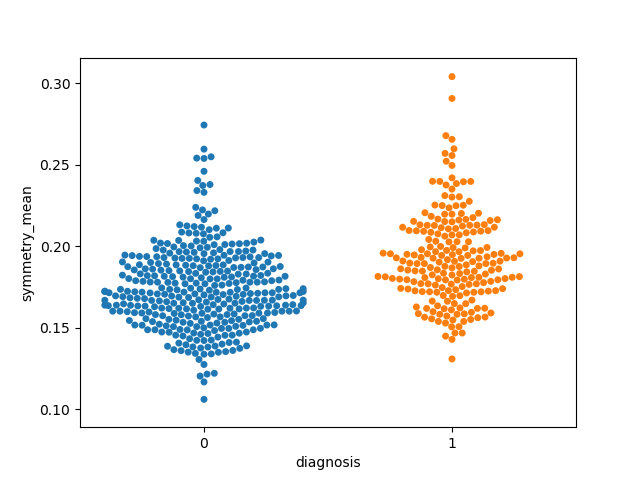}}\\
\subfloat[][Diagnosis vs fractal mean]{\includegraphics[width=0.3\textwidth]{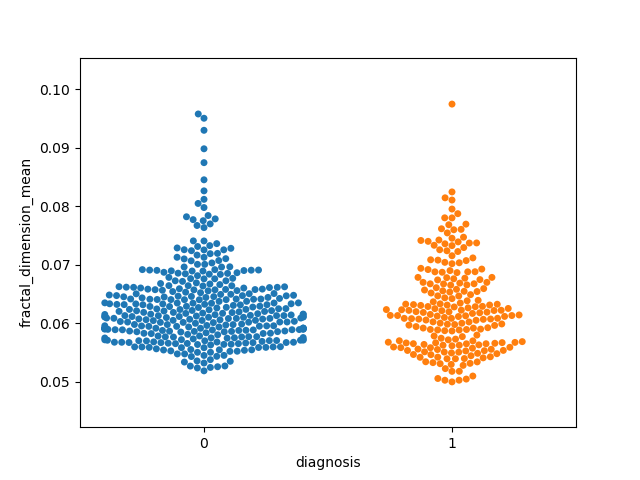}}
\caption{Wisconsin Diagnosis Breast Cancers Data visualization}
\end{figure*}

The WDBC dataset is used to exploit the machine learning algorithms. We have conducted this rigorous experiment in two phases which is listed as follows-
\begin{itemize}
\item \textbf{Phase I:} We input the original WDBC dataset to the machine learning algorithms for 100 times and results are plotted in the chart. 
\item \textbf{Phase II:} We double the WDBC dataset by duplicating the dataset and input to the machine learning algorithms. The outcome of the experiments is plotted in chart.
\end{itemize}

\section{Experimentations and Results}
\label{er}
\begin{figure*}[ht]
\centering
\includegraphics[width=0.95\textwidth]{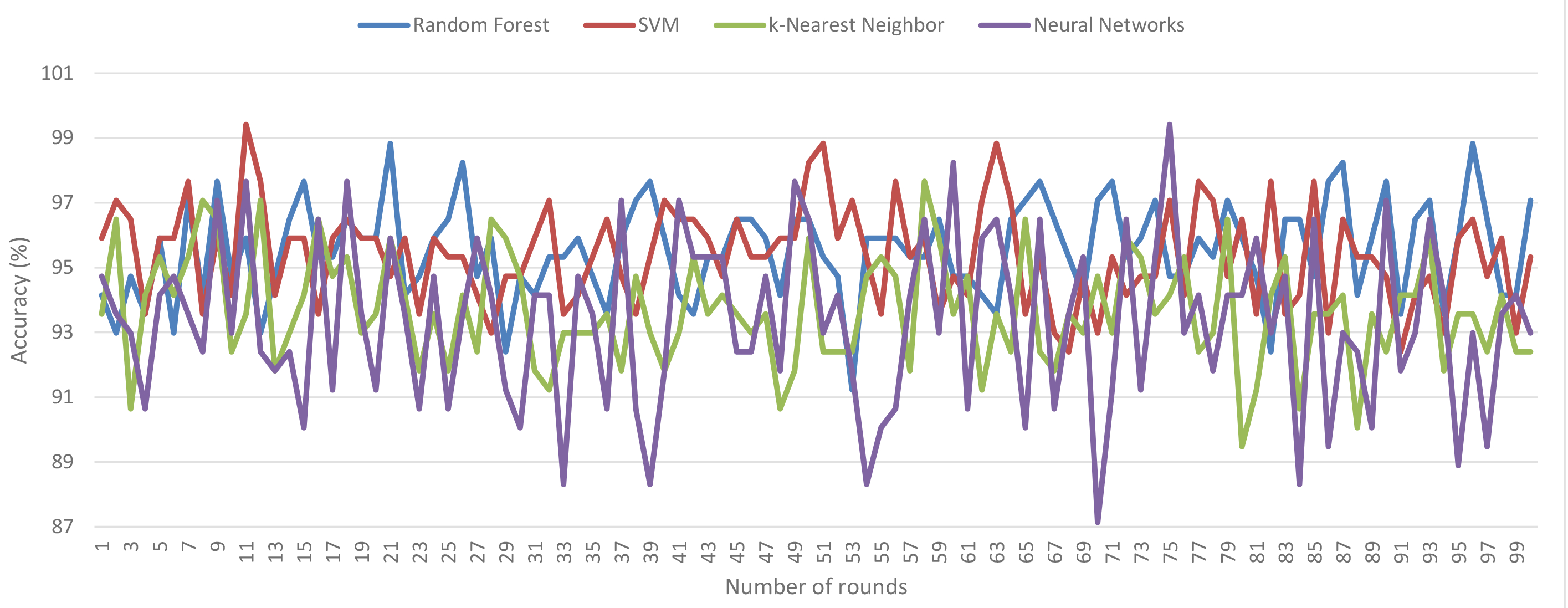}
\caption{The statistics of Random Forest, SVM, k-Nearest Neighbor, and Neural Networks in 100 rounds with original dataset.}
\label{acc1}
\end{figure*}

\begin{figure*}[ht]
\centering
\includegraphics[width=0.95\textwidth]{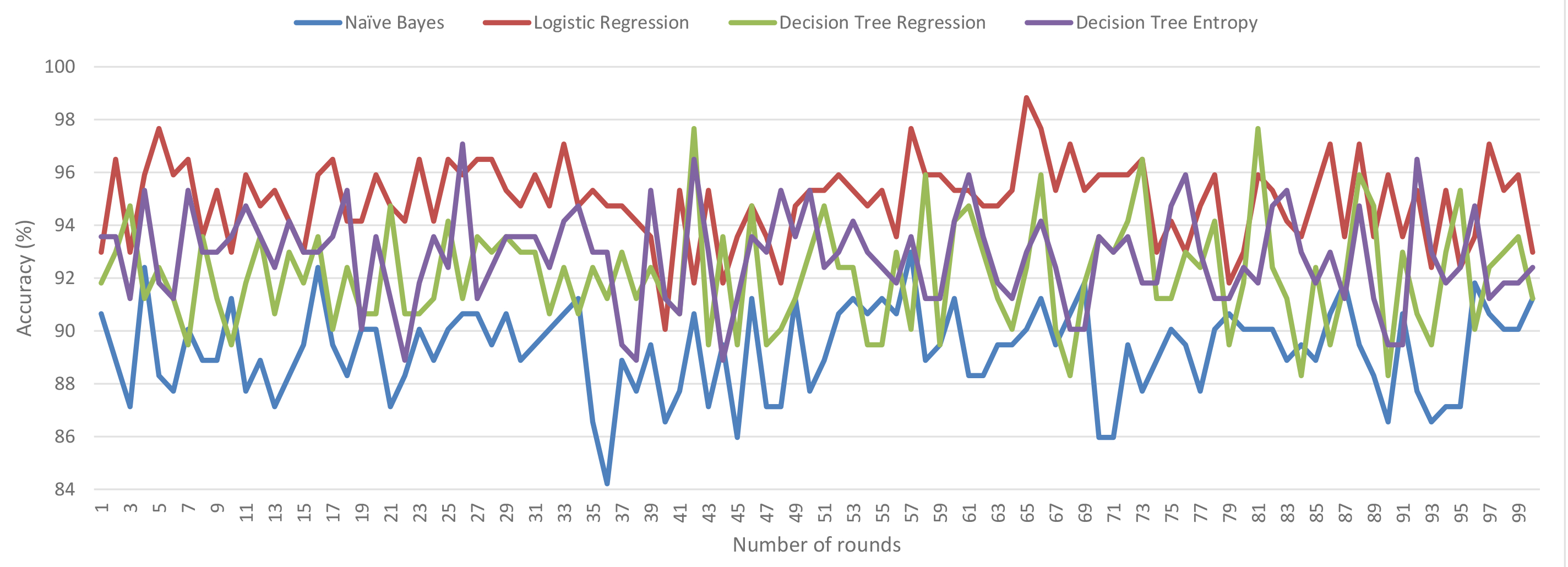}
\caption{The statistics of Naive Bayes, Logistic Regression, Decision Tree entropy, and Decision Tree regressor in 100 rounds with original dataset.}
\label{acc2}
\end{figure*}

Figure~\ref{acc1} depicts the 100 rounds prediction accuracy of Random Forest, SVM, k-Nearest Neighbor, and Neural Networks. Similarly, Figure~\ref{acc2} depicts the 100 rounds prediction accuracy of Naive Bayes, Logistic Regression, Decision Tree entropy, and Decision Tree regressor. Overall, the Random Forest model performs excellent in prediction and Naive Bayes performs very poorly during the 100 round iterations. As per our experienced, the Neural Network takes huge time in training and testing.    

Figure~\ref{acc3}, \ref{acc4} and ~\ref{acc5} depicts the best, average and worst case accuracy in prediction of Random Forest, Support Vector Machine (SVM), k-Nearest Neighbor, Neural Networks, Naive Bayes, Decision Tree entropy, and Decision Tree regressor. The worst performer is Naive Bayes algorithm in this dataset. The best, average and worst case prediction accuracy of Naive Bayes are $92.98245614$, $89.28654971$, and  $84.21052632$ respectively. The highest 'Best Case' is achieved by SVM and  Neural Networks which is 99.41520468 for both. The Random Forest model consistently predicts with high accuracy on an average $95.5380117$. It outperforms SVM and all other learning models as shown in Figure~\ref{acc3}, \ref{acc4} and ~\ref{acc5}. The Best Case of Random Forest is slightly lower than SVM and Neural Network. However, the SVM is the best in Best Case (99.41520468) and Worst Case (92.39766082). The Neural Network shows poor performance in Worst case (87.13450292) and average case (93.32748538). The accuracy of the learning models fluctuates with the times due to random samples taken from the input. Therefore, we perform 100 times experiments with the same dataset to extract the mean value. 

The randomness in Machine Learning algorithms makes difficult to decide the prediction accuracy. As per our experience, the prediction accuracy always varies. Figure ~\ref{acc1}, and \ref{acc2} depicts the randomness of the accuracy during 100 rounds training and testing. We have observed that the results never remain same as previous result. Therefore, it is very easy to claim the maximum accuracy with highest possible results which cannot be validated or believed easily. Is it wiser way to believe others research results? Or can a researcher manipulate their results? These questions arrives in randomness.

\subsection{Accuracy measurement with various partition}
\begin{figure}[ht]
\centering
\includegraphics[width=0.45\textwidth]{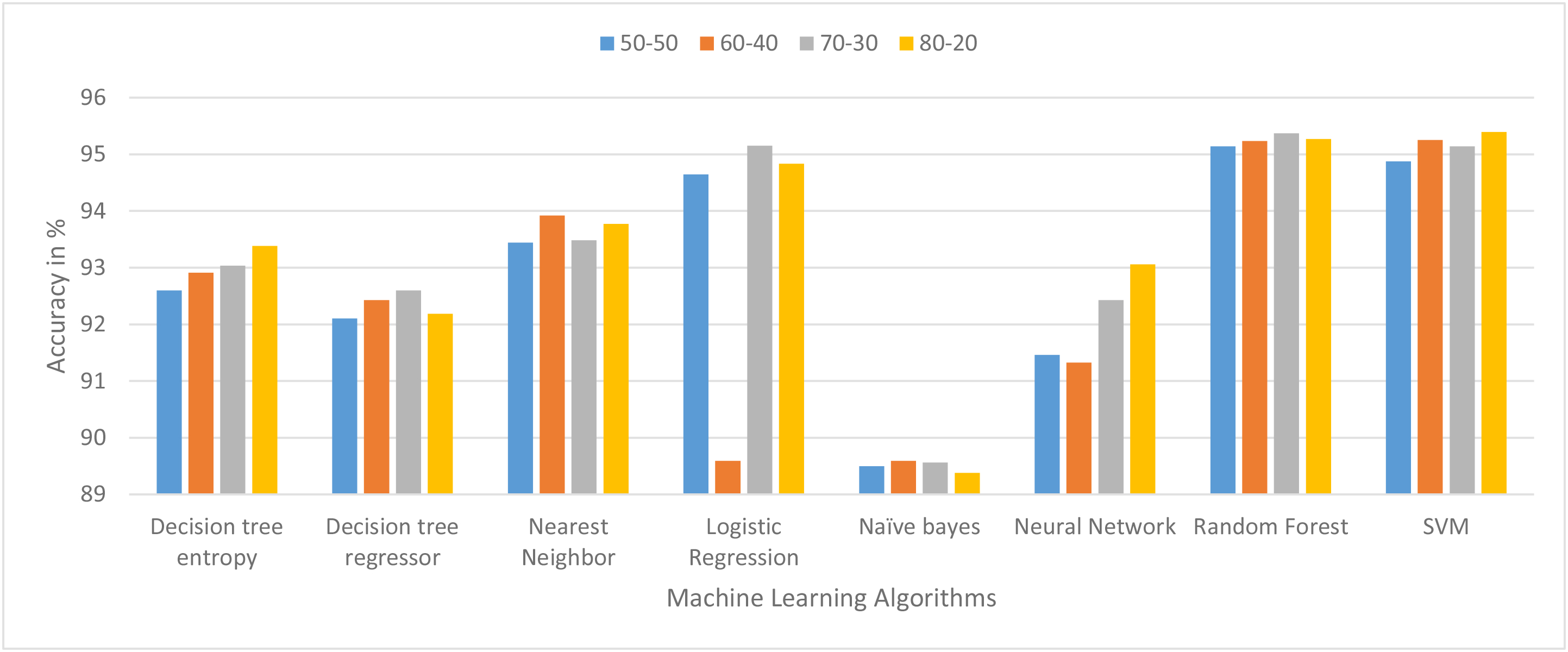}
\caption{Average accuracy calculation of some machine learning algorithms on UCI Breast Cancer dataset. Considering every possible split as 50-50, 60-40, 70-30, and 80-20.}
\label{acc3}
\end{figure}

Figure ~\ref{acc3} shows the average accuracy calculation on various machine learning algorithms. The dataset is split into 50-50, 60-40, 70-30, and 80-20. In this case, we have observed that 80-20 accuracy is better than other splitting. Moreover, most of the article reports 70-30 and it is assumed as standard practice. The SVM and Random Forest algorithm excel all the split. Logistic regression outperform Nearest Neighbor and Neural Network. However, the Logistic regression exhibits its poor performance in 60-40 split.

\begin{figure}[!ht]
\centering
\includegraphics[width=0.45\textwidth]{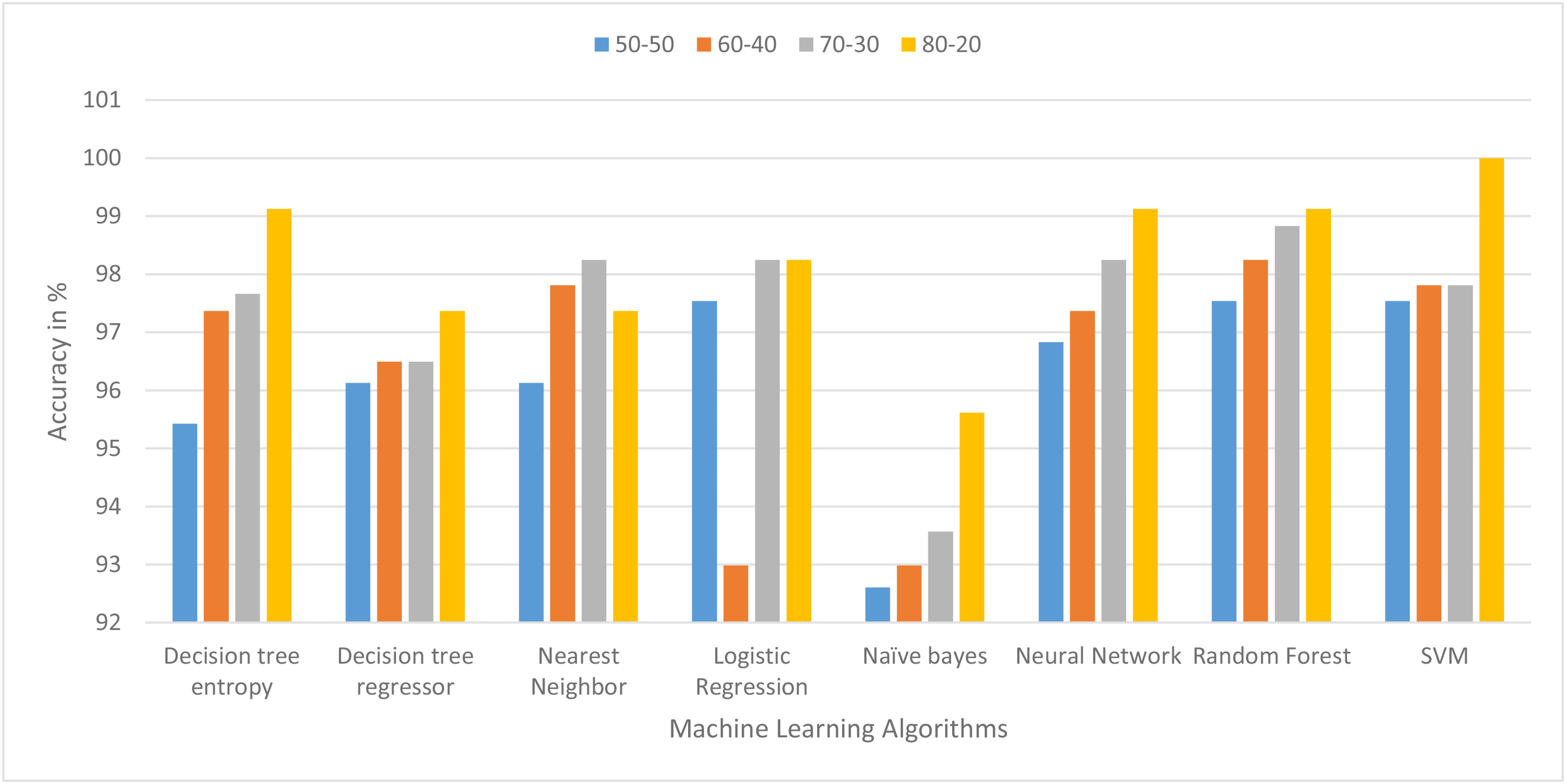}
\caption{Maximum accuracy calculation of some machine learning algorithms on UCI Breast Cancer dataset. Considering every possible split as 50-50, 60-40, 70-30, and 80-20.}
\label{acc4}
\end{figure}

Figure~\ref{acc4} shows the maximum accuracy in 50-50, 60-40, 70-30, and 80-20 split. The SVM achieves highest accuracy in 80-20 dataset split. The Random Forest performs better than all other algorithms in 50-50, 60-40, and 70-30 dataset split. However, the maximum does not count in life threatening diseases.

\begin{figure}[!ht]
\centering
\includegraphics[width=0.45\textwidth]{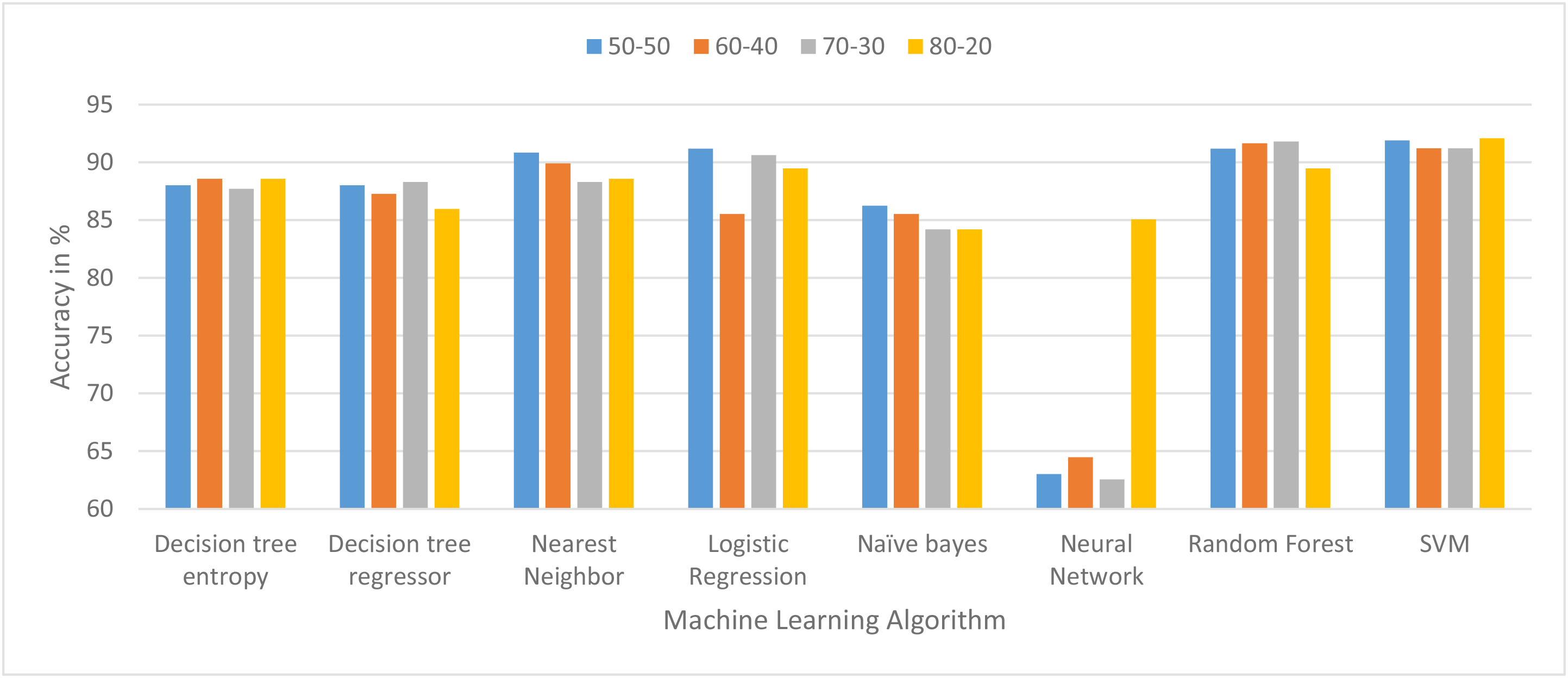}
\caption{Minimum accuracy calculation of some machine learning algorithms on UCI Breast Cancer dataset. Considering every possible split as 50-50, 60-40, 70-30, and 80-20.}
\label{acc5}
\end{figure}

Figure~\ref{acc3}, \ref{acc4}, and \ref{acc5} shows the mean accuracy, maximum accuracy, and minimum accuracy respectively. The figures show the accuracy of Naive Bayes, Nearest Neighbor, Neural Network, Decision Tree Entropy, Decision Tree Regressor, Logistic Regression, Random Forest and SVM. SVM performs the best in 50-50 and 80-20 split and Random Forest perform the best in 60-40 and 70-30 split in minimum accuracy. Nearest Neighbor performs well in 50-50 split and Logistic regression perform well in 50-50 split in the case of minimum accuracy. Surprisingly, the Neural Network performs worst in Minimum accuracy in all split. Therefore, we cannot rely on Neural Network algorithm, albeit the algorithm gives the best result in some cases. However, in this evaluation, the Random forest outperforms all other algorithms.

\subsection{Doubling the dataset}

\begin{figure}[!ht]
\centering
\includegraphics[width=0.45\textwidth]{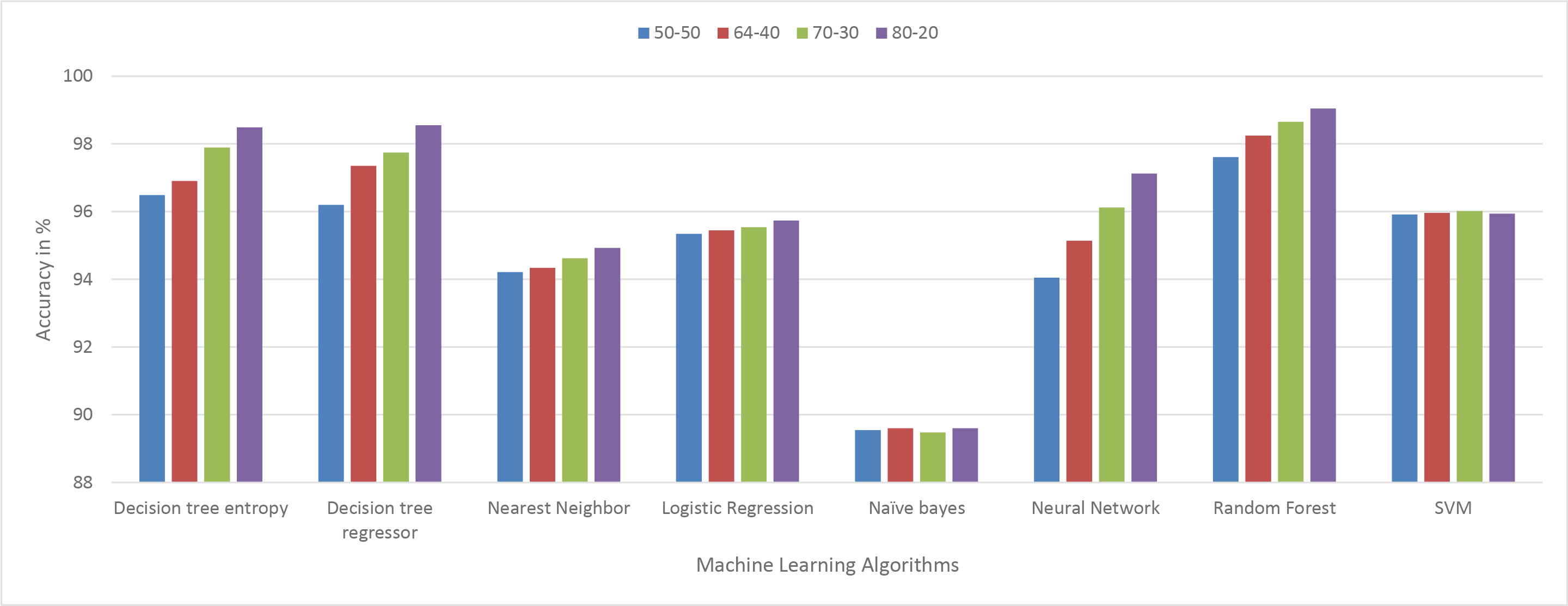}
\caption{Average accuracy calculation of some machine learning algorithms on UCI Breast Cancer dataset by doubling the dataset. Considering every possible split as 50-50, 60-40, 70-30, and 80-20.}
\label{acc6}
\end{figure}

\begin{figure}[!ht]
\centering
\includegraphics[width=0.45\textwidth]{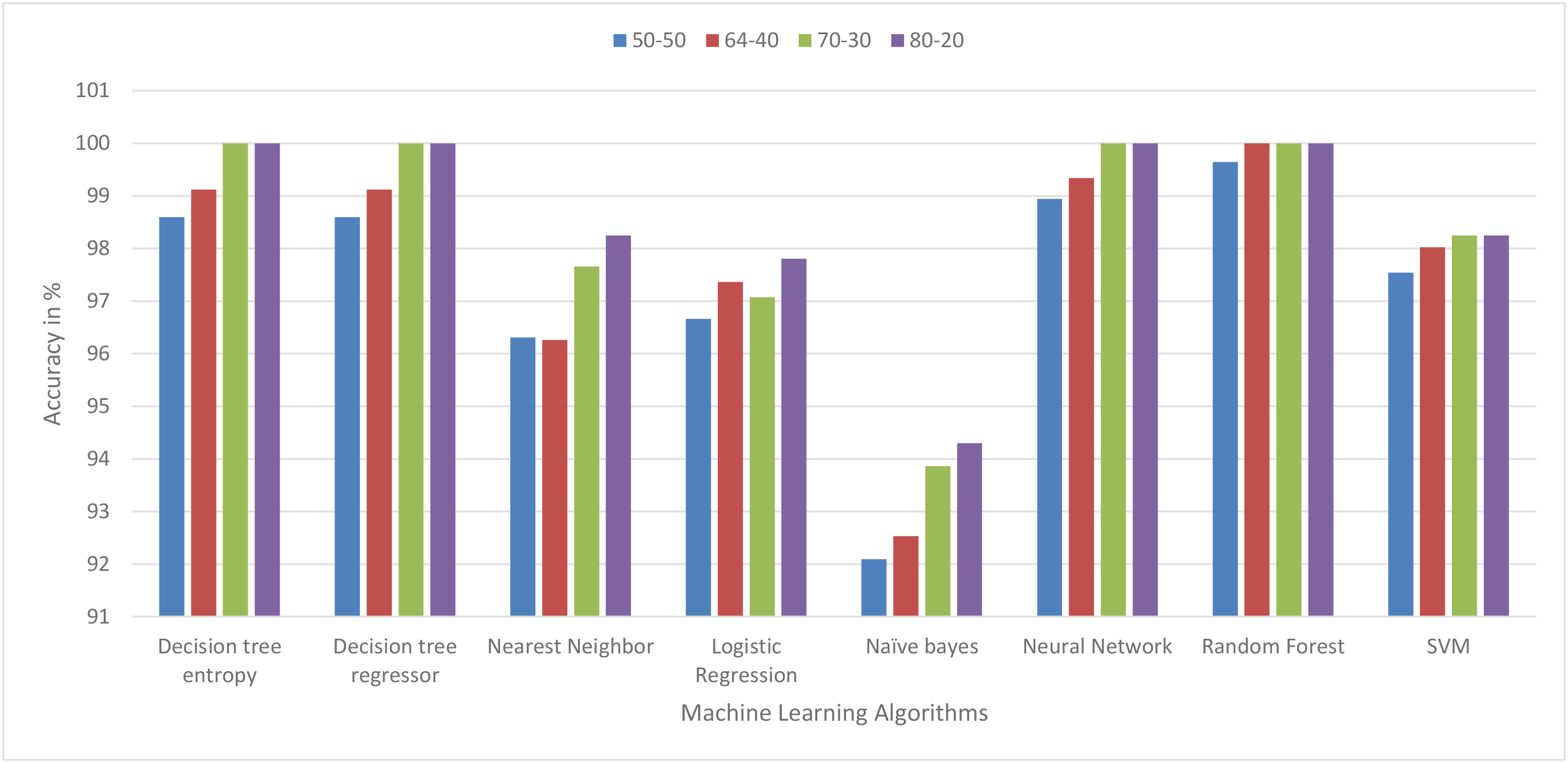}
\caption{Maximum accuracy calculation of some machine learning algorithms on UCI Breast Cancer dataset by doubling the dataset. Considering every possible split as 50-50, 60-40, 70-30, and 80-20.}
\label{acc7}
\end{figure}

\begin{figure}[!ht]
\centering
\includegraphics[width=0.45\textwidth]{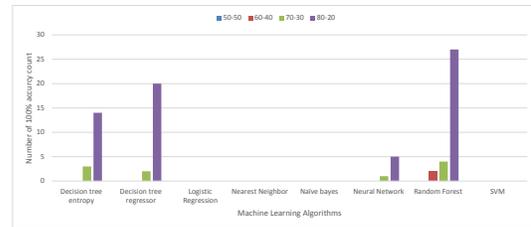}
\caption{Number of hundred percent accuracy counts in machine learning algorithms during 100 runs}
\end{figure}

A 100\% accuracy is unbelievable! It's surprising! However, we have achieved. The Random Forest shows the maximum accuracy of 100\% in the Wisconsin Breast Cancer dataset by doubling the input size.  The accuracy of Random Forest, Neural Network, and Decision Tree reached to 100\% in the best case by doubling the input dataset. The Nearest Neighbor, Naive Bayes, and SVM could not reach to hundred percent. However, the accuracy also increased in this input. The total hundred percent accuracy count is maximum in 80-20 partition and the Random Forest exhibit the highest of 21s hundred count in 100 runs. Moreover, the average accuracy of all algorithms has raised. Random Forest shows the best performance in the average case. And, the decision tree also performs satisfactory in the average case. 

\begin{figure}[!ht]
\centering
\includegraphics[width=0.45\textwidth]{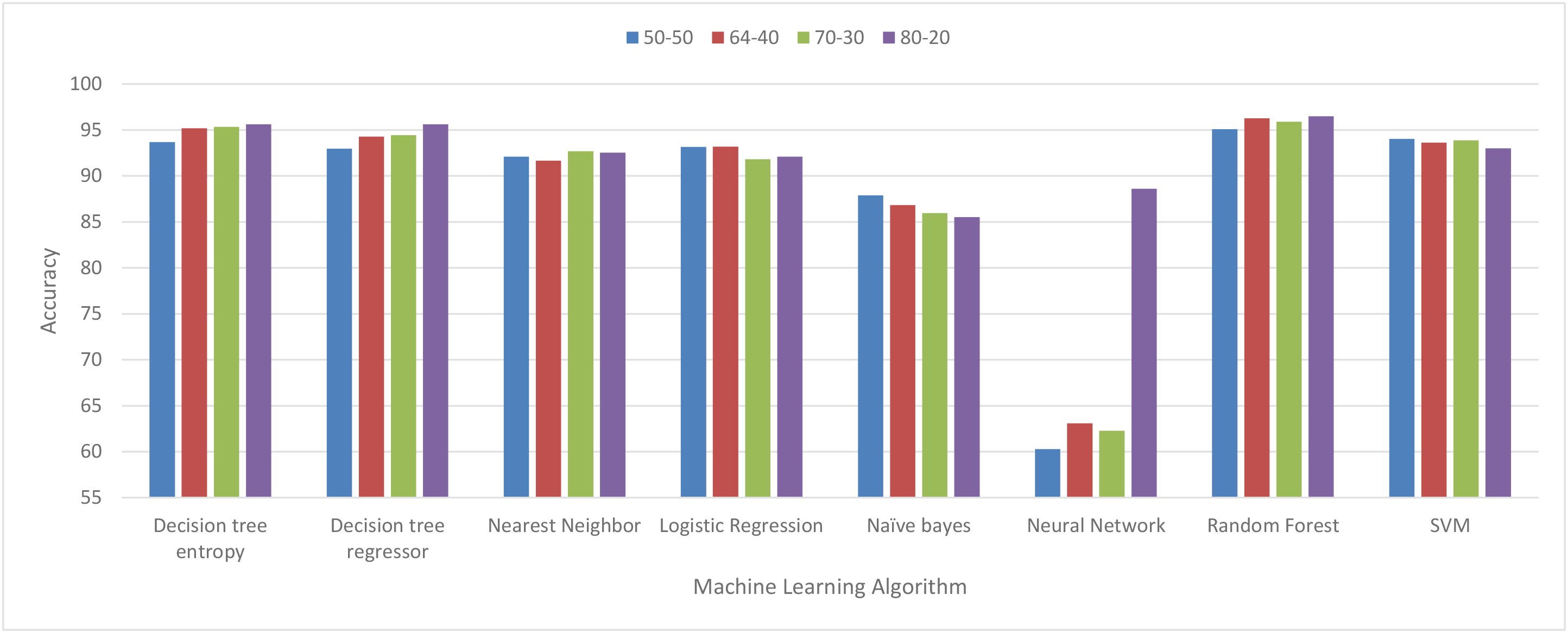}
\caption{Minimum accuracy calculation of some machine learning algorithms on UCI Breast Cancer dataset by doubling the dataset. Considering every possible split as 50-50, 60-40, 70-30, and 80-20.}
\label{acc8}
\end{figure}

The Random Forest excels in prediction in minimum accuracy, however, the all other algorithms also performs well except Neural Network. The result shows that the accuracy of a machine learning algorithm can easily be manipulated. The Random Forest model is more vulnerable to this kind of malicious result intentionally or unintentionally. However, the SVM is not affected more with the doubling the size, but we also observed the low rises in accuracy. 

\section{Discussion}
\label{dis}
\begin{table}[!ht]
\centering
\caption{Misclassification of one and its consequences}
\begin{tabular}{ccc}
\hline
\textbf{Input} & \textbf{Misclassification} & \textbf{Accuracy}\\ \hline
2 & 1 & $50\%$ \\
5 & 1 & $80\%$ \\
10 & 1 & $90\%$\\
20 & 1 & $95\%$\\
30 & 1 & $96\%$\\
40 & 1 & $97.5\%$\\
50 & 1 & $98\%$\\
60 & 1 & $98.33\%$\\
70 & 1 & $98.57\%$\\
80 & 1 & $98.75\%$\\
90 & 1 & $98.89\%$\\
100 & 1 & $99\%$\\
\hline

\end{tabular}
\label{tab}
\end{table}
Table ~\ref{tab} exposes one of the reasons for increasing accuracy. An increment of input increases the accuracy. The machine learning algorithm depends on the input size. An input of 2 and one misclassification causes 50\% degradation of accuracy. While same number of misclassification in 100 inputs causes 99\% accuracy. The accuracy also increased in doubling the data. Moreover, the machine learning algorithms selects a random sample from the input which is more accurate to predict. Because, the data sample is duplicated and one of the samples is picked and matched with a duplicate sample data. Thus, accuracy increases to 100\%. However, this practice is unethical. On the contrary, a large amount of data can be generated using genetic algorithm. A malignant and a benign dataset can be used to generate offspring randomly by crossover method. This is not deployable in real life, however, we can evaluate experimentally the performance of the machine learning algorithm by the large set of dataset.

\section{Conclusion}
\label{con}
As we have shown that we have achieved 100\% accuracy at maximum. We illustrate that maximum accuracy is not a significant factor in life threatening diseases. The minimum accuracy plays utmost important in benchmarking process and real life scenario in the case of life threatening diseases, for instance, Cancer. The paper also discusses on how to achieve 100\% accuracy, using machine learning algorithm. Also, we demonstrated the unethical way of reporting accuracy of machine learning algorithm which can easily mislead the algorithm intentionally or unintentionally. Enhancing the maximum accuracy does not impact in Cancer Computing. On the contrary, most of the researchers interested in the enhancement of maximum accuracy which does not serve the purpose of cancer computing.

%\subsection{References}
\bibliographystyle{IEEEtran}
\bibliography{mybibfile}

\begin{thebibliography}{10}
\providecommand{\url}[1]{#1}
\csname url@rmstyle\endcsname
\providecommand{\newblock}{\relax}
\providecommand{\bibinfo}[2]{#2}
\providecommand\BIBentrySTDinterwordspacing{\spaceskip=0pt\relax}
\providecommand\BIBentryALTinterwordstretchfactor{4}
\providecommand\BIBentryALTinterwordspacing{\spaceskip=\fontdimen2\font plus
\BIBentryALTinterwordstretchfactor\fontdimen3\font minus
  \fontdimen4\font\relax}
\providecommand\BIBforeignlanguage[2]{{%
\expandafter\ifx\csname l@#1\endcsname\relax
\typeout{** WARNING: IEEEtran.bst: No hyphenation pattern has been}%
\typeout{** loaded for the language `#1'. Using the pattern for}%
\typeout{** the default language instead.}%
\else
\language=\csname l@#1\endcsname
\fi
#2}}

\bibitem{WHO}
WHO, ``Cancer fact sheet 2018,'' Accessed on 20 March 2018 from
  \url{http://www.who.int/mediacentre/factsheets/fs297/en/}.

\bibitem{Chen2014}
Y.-C. Chen, W.-C. Ke, and H.-W. Chiu, ``Risk classification of cancer survival
  using ann with gene expression data from multiple laboratories,''
  \emph{Computers in Biology and Medicine}, vol.~48, pp. 1 -- 7, 2014.

\bibitem{xu}
X.~Xu, Y.~Zhang, L.~Zou, M.~Wang, and A.~Li, ``A gene signature for breast
  cancer prognosis using support vector machine,'' in \emph{2012 5th
  International Conference on BioMedical Engineering and Informatics}, 2012,
  pp. 928--931.

\bibitem{Exa}
K.~P. Exarchos, Y.~Goletsis, and D.~I. Fotiadis, ``Multiparametric decision
  support system for the prediction of oral cancer reoccurrence,'' \emph{IEEE
  Transactions on Information Technology in Biomedicine}, vol.~16, no.~6, pp.
  1127--1134, 2012.

\bibitem{Ahmad}
P.~A. E. M. R.~A. Ahmad~LG, Eshlaghy~AT, ``Using three machine learning
  techniques for predicting breast cancer recurrence,'' \emph{J Health Med
  Inform}, vol.~4, no. 124, 2013.

\bibitem{Glob}
GLOBOCAN, ``{GLOBOCAN 2012}: Estimated cancer incidence, and prevalence
  worldwide in 2012,'' Accessed on 20 March 2018 from
  \url{http://globocan.iarc.fr/Pages/fact\_sheets\_cancer.aspx}.

\bibitem{Jhajharia}
S.~Jhajharia, H.~K. Varshney, S.~Verma, and R.~Kumar, ``A neural network based
  breast cancer prognosis model with pca processed features,'' in \emph{2016
  International Conference on Advances in Computing, Communications and
  Informatics (ICACCI)}, Sept 2016, pp. 1896--1901.

\bibitem{Kourou}
K.~Kourou, T.~P. Exarchos, K.~P. Exarchos, M.~V. Karamouzis, and D.~I.
  Fotiadis, ``Machine learning applications in cancer prognosis and
  prediction,'' \emph{Computational and Structural Biotechnology Journal},
  vol.~13, pp. 8 -- 17, 2015.

\bibitem{Luke}
L.~Barracliffe, O.~Arandelovic, and G.~Humphris, \emph{A pilot study of breast
  cancer patients: Can machine learning predict healthcare professionals'
  responses to patient emotions?}\hskip 1em plus 0.5em minus 0.4em\relax
  International Society for Computers and Their Applications, 2017, pp.
  101--106.

\bibitem{Kesler}
S.~R. Kesler, A.~Rao, D.~W. Blayney, I.~A. Oakley-Girvan, M.~Karuturi, and
  O.~Palesh, ``Predicting long-term cognitive outcome following breast cancer
  with pre-treatment resting state fmri and random forest machine learning,''
  \emph{Frontiers in Human Neuroscience}, vol.~11, p. 555, 2017.

\bibitem{Delen}
D.~Delen, G.~Walker, and A.~Kadam, ``Predicting breast cancer survivability: a
  comparison of three data mining methods,'' \emph{Artificial Intelligence in
  Medicine}, vol.~34, no.~2, pp. 113 -- 127, 2005.

\bibitem{Hamsagayathri}
P.~Hamsagayathri and P.~Sampath, ``Priority based decision tree classifier for
  breast cancer detection,'' in \emph{2017 4th International Conference on
  Advanced Computing and Communication Systems (ICACCS)}, Jan 2017, pp. 1--6.

\bibitem{Nguyen}
Y.~W. Cuong~Nguyen and H.~N. Nguyen, ``Random forest classifier combined with
  feature selection for breast cancer diagnosis and prognostic,'' \emph{J.
  Biomedical Science and Engineering}, 2013.

\end{thebibliography}

\end{document}